\documentclass[lettersize,journal]{IEEEtran}

\usepackage{flushend}
\usepackage{tabularx}
\usepackage{amssymb}

\newcolumntype{D}{>{\centering\arraybackslash}X}
\newcolumntype{C}{>{\hsize=0.5\hsize}D}

\usepackage[caption=false,font=normalsize,labelfont=sf,textfont=sf]{subfig}
\usepackage{graphicx}
\usepackage{standalone}
\usepackage{tikz}
\usetikzlibrary{shapes,backgrounds,shapes.geometric}
\usepackage{circuitikz}
\tikzset{font={\fontsize{9pt}{9}\selectfont}}
\usepackage{pgfplots}
\usepgfplotslibrary{fillbetween,colormaps,groupplots,statistics}
\pgfplotsset{compat=newest}

\usepackage{tabularx}
\usepackage{makecell}
\usepackage{booktabs}
\usepackage{multirow}
\newcolumntype{Y}{>{\centering\arraybackslash}X}
\newcolumntype{s}{>{\hsize=0.7\hsize}Y}

\usepackage{siunitx}
\usepackage{amsmath}
\usepackage{amsfonts}
\usepackage{scalerel}
\usepackage{amssymb}
\usepackage{bm}

\usepackage{algorithm}
\usepackage{algorithmicx}

\usepackage{cite}
\usepackage{jabbrv}

\usepackage[utf8]{inputenc}
\usepackage{enumitem}

\date{\today}

\usepackage{hyperref}
\hypersetup{
	pdfauthor={Julian Schumann},
	pdfcreator={Julian Schumann},
	pdfproducer={},
	pdfsubject={First project for PhD},
    colorlinks={true},
    linkcolor={blue},
    citecolor={blue},
    urlcolor={red!50!white},
    hyperindex={true}
}
\makeatletter
\hypersetup{pdftitle=\@title}
\makeatother
\usetikzlibrary{external}
\title{STEP: Structured Training and Evaluation Platform for benchmarking trajectory prediction models}
\author{Julian F. Schumann$^{*}$, Anna M\'esz\'aros$^{*}$, Jens Kober, Arkady Zgonnikov
\thanks{$^*$Shared first authorship; The authors are with the Department of Cognitive Robotics, Delft University of Technology, Delft, Zuid Holland 2628 CD, The Netherlands (e-mail: j.f.schumann@tudelft.nl; a.meszaros@tudelft.nl; j.kober@tudelft.nl; a.zgonnikov@tudelft.nl) \textit{(Corresponding Author: Julian Schumann)}}
\thanks{The source code, trained models, and data can be found online at  \href{https://github.com/DAI-Lab-HERALD/General-Framework}{public Github repository}}}

\begin{document}
\twocolumn
\maketitle

\begin{abstract}
    While trajectory prediction plays a critical role in enabling safe and effective path-planning in automated vehicles, standardized practices for evaluating such models remain underdeveloped. Recent efforts have aimed to unify dataset formats and model interfaces for easier comparisons, yet existing frameworks often fall short in supporting heterogeneous traffic scenarios, joint prediction models, or user documentation. In this work, we introduce STEP -- a new benchmarking framework that addresses these limitations by providing a unified interface for multiple datasets, enforcing consistent training and evaluation conditions, and supporting a wide range of prediction models. We demonstrate the capabilities of STEP in a number of experiments which reveal 1) the limitations of widely-used testing procedures, 2) the importance of joint modeling of agents for better predictions of interactions, and 3) the vulnerability of current state-of-the-art models against both distribution shifts and targeted attacks by adversarial agents. With STEP, we aim to shift the focus from the ``leaderboard'' approach to deeper insights about model behavior and generalization in complex multi-agent settings.
\end{abstract}
\begin{IEEEkeywords}
autonomous driving, modelling and simulation, human behavior prediction, benchmarking.
\end{IEEEkeywords}
\section{Introduction}
Modern automated vehicle (AV) systems are typically organized into modular software stacks, consisting out of components such as perception, prediction, and planning. This modular structure allows for an independent development and replacement of these components, enabling greater transparency and scalability. The development of new prediction modules has been receiving increased interest in the research community~\cite{mozaffari_deep_2022, sun2020scalability}, with many attempts at improving the accuracy and reliability of predictions. 

Rigorous evaluation of prediction models is paramount for ensuring reliable AV performance.
However, in the current literature there is a lack of established benchmarks for trajectory prediction models. Despite (or possibly due to) the increasing availability of diverse large-scale datasets, there are no widely accepted  evaluation approaches and guidelines. This results in two practical issues that model developers face. 
First, besides implementing a new model, the developers currently need to implement model evaluation from scratch, such as the coding of data-loaders providing both training and testing data in the format their model requires. This introduces much overhead during the model development.
Second, the lack of established routines can lead to splintered evaluation, preventing equitable comparisons with state-of-the-art models (especially if they use different data interfaces). For instance, the calculation of performance metrics used for model comparison can differ from paper to paper (e.g., with respect to which parts of the dataset are included), while even slight differences can have major impact on the reported results. 

Motivated by these challenges, recent work has taken steps towards unified evaluation of trajectory prediction models, such as a unified interface to different datasets~\cite{ivanovic_trajdata_2023} and frameworks allowing more streamlined model comparisons~\cite{kothari2021human, rudenko2022atlas, schumann_benchmarking_2023, feng_unitraj_2024}. However, such frameworks still lack many features that are essential for a holistic and fair comparison of trajectory prediction models. \emph{Trajdata}~\cite{ivanovic_trajdata_2023}, for example, only provides a unified interface between datasets, with no way to ensure that all models are evaluated under the same conditions.
TrajNet++~\cite{kothari2021human} and ATLAS~\cite{rudenko2022atlas}, meanwhile, made first steps towards developing a complete benchmarking framework for trajectory prediction. However, these approaches are not widely applicable, as they only allow pedestrians. Correspondingly, there is also a neglect of features more relevant in heterogeneous traffic, such as lane information and agent information like type and size. The gap acceptance benchmarking framework (GAP)~\cite{schumann_benchmarking_2023}, being designed to also evaluate models predicting different classes of behaviors, allows evaluation beyond just predicted trajectories. However, it is limited to gap acceptance scenarios only, and thus unable to evaluate models in other scenarios common in large datasets~\cite{sun2020scalability,wilson2021argoverse}.
\emph{UniTraj}~\cite{feng_unitraj_2024} addresses many of the limitations mentioned above, but still lacks the support for models that predict the future trajectory of all agents in a scene jointly and simultaneously (which is the case for many state-of-the-art models~\cite{aydemir_adapt_2023,rowe_fjmp_2023,girgis_latent_2022,shi_mtr_2024}), as well as out-of-the-box evaluation of model robustness against adversarial attacks, which is an increasingly important consideration for trajectory prediction~\cite{hagenus2024survey}. Importantly, existing benchmarks also lack supporting documentation for the addition of new datasets or models, resulting in a high entry barrier for model developers. 

\begin{table*}[]
    \centering
    \caption{Comparison between existing benchmarks for trajectory prediction. An asterisk $^*$ next to a checkmark (\checkmark) indicates that the feature is available to a limited extent. The letters for agent type correspond to the following -- B: Bicycle, M: Motorcycle, P: Pedestrian, V: Vehicle.}
    \label{tab:frameworkComparison}
    \begin{tabularx}{\textwidth}{|X|C|C|C|C|C|C|}
    \hline
      Framework   & trajdata~\cite{ivanovic_trajdata_2023} & TrajNet++~\cite{kothari2021human} & ATLAS~\cite{rudenko2022atlas} & UniTraj~\cite{feng_unitraj_2024} & GAP~\cite{schumann_benchmarking_2023} & STEP (Ours) \\ \hline
      Unified data format & \checkmark & \checkmark & \checkmark & \checkmark & \checkmark & \checkmark \\
      Large datasets supported & \checkmark &  &  & \checkmark & & \checkmark \\
      Evaluation with unified metrics & & \checkmark & \checkmark & \checkmark & \checkmark & \checkmark \\
      Controllable dataset splitting & & & & \checkmark$^*$ & \checkmark & \checkmark \\
      Environment information & image \& graph & & image & image \& graph & image & image \& graph\\
      Additional agent data & type \& size &  &  & type \& size & type & type \& size\\
      Included agent types & B, M, P \& V & P & P & B, P, V & P \& V & B, M, P \& V\\
      Adjustable input/output length & \checkmark & & \checkmark & \checkmark & \checkmark & \checkmark \\
      Adjustable trajectory frequency & \checkmark &  & \checkmark & & \checkmark & \checkmark  \\
      Supporting scene-level models & \checkmark & & & & & \checkmark \\
      Includes behavior classification & & \checkmark$^*$ & &  \checkmark$^*$ & \checkmark & \checkmark\\
      Validation of model's robustness & & & \checkmark$^*$ & & & \checkmark \\
      Documentation supporting addition of new modules & \checkmark & & \checkmark$^*$ & \checkmark$^*$ & & \checkmark  \\ \hline
    \end{tabularx}
\end{table*}

To address these issues, this paper introduces STEP (Structured Training and Evaluation Platform), which provides a unified interface across numerous datasets and enabling flexible formatting of the data to create diverse testing scenarios. STEP enables comparisons with a broad selection of current state-of-the-art models, including single agent trajectory prediction models~\cite{salzmann_trajectron_2022, nayakanti_wayformer_2023, scholler_flomo_2021, meszaros_trajflow_2024} and scene-level joint trajectory prediction models~\cite{yuan_agentformer_2021, aydemir_adapt_2023, rowe_fjmp_2023, girgis_latent_2022}. Importantly, STEP ensures that all evaluated models are trained and tested under the same conditions, making those comparisons truly equitable.
The framework is supported by a detailed documentation, enabling easy addition of new datasets, models, and metrics.

\section{Related Work}
In Table~\ref{tab:frameworkComparison} we provide a high-level overview of the features of existing benchmarking frameworks, highlighting the added functionalities of our approach.
First, STEP is the only approach that provides meaningful flexibility over the mechanism used for dividing a dataset into the training and testing subsets.
For example, the drone datasets (such as highD~\cite{krajewski_highd_2018} and rounD~\cite{krajewski_round_2020}) have recordings from multiple locations. Practitioners who use these datasets split them in multiple different ways, such as following an N-cross-validations paradigm across the whole dataset~\cite{meszaros_trajflow_2024}, or taking a fraction of the recordings from each location~\cite{diehl2019graph}, or simply via a single random split~\cite{wen2022social, mozaffari2023trajectory, schumann_using_2023}.
Alternatively, one may wish to split the data by the specific locations following a leave-one-out paradigm~\cite{gupta_social_2018} (often used for the ETH/UCY dataset~\cite{pellegrini2009you, lerner_crowds_2007}) where all but one location is used for training and the held out location is used for testing. Among existing frameworks, only STEP provides all those functionalities without the need to implement them from scratch.

Second, model developers usually evaluate models on multiple different datasets, but typically training and testing subsets belong to the same dataset. At the same time, combining different datasets together for training and testing can provide more generalized evaluation~\cite{feng_unitraj_2024}. While this is theoretically possible in UniTraj~\cite{feng_unitraj_2024}, one would still only test the model on the part of one selected dataset pre-labeled as testing data. STEP supports and generalizes this approach by giving the user a large number of out-of-the-box options for splitting composite datasets, such as testing only on specific locations picked from any of the composite datasets.

Third, the existing frameworks give the user only limited control over the observation horizon and especially the frequency of the input. Previous research has shown that the duration of the input can play an important role in model performance~\cite{monti2022many, xu2024adapting}; similarly, the frequency of the input might also affect performance.
Additionally, keeping in mind the bigger picture of the AV stack and the fact that different perception modules may operate at different frequencies, it would be beneficial for model developers if they could train their models already accounting for the operational frequency of the upstream perception module. However, of the existing frameworks, only ATLAS~\cite{rudenko2022atlas} give users control over both duration and frequency of the input. STEP supports this but in addition also allows the user to ensure that varying the input data does not change the selection of included scenario, guaranteeing that differences in performance actually stem from the differences in the input data and not from corresponding changes in the size or structure of the dataset.

Fourth, scene-level predictions, i.e. joint predictions for all agents in the scene, are receiving increasing  attention~\cite{rowe_fjmp_2023,girgis_latent_2022,aydemir_adapt_2023, shi_mtr_2024}. The reason for this is that single-agent prediction models are unequipped to properly account for future interactions between agents. To facilitate the development of such models, the evaluation framework must not only provide past and future information of all relevant agents in a scene during training, but also make it possible to evaluate those joint predictions together. 
Furthermore, such joint prediction models can profit from the use of interaction-based metrics (i.e., metrics accounting for the scene-level predictions). However, only TrajNet++~\cite{kothari2021human} provides interaction-based metrics, but limits them to two-agent interactions and does not account for the complete scene-level distribution.
As such, none of the existing frameworks are suitable for developing scene-level (joint) prediction models, while STEP fully supports these.

Finally, model robustness is becoming increasingly important within the field~\cite{hagenus2024survey}. While ATLAS~\cite{rudenko2022atlas} can test the robustness of models against white noise, this only represents a small part of potential perturbations of input data. Specifically, targeted attacks by adversarial agents~\cite{cao_advdo_2022, zhang2022adversarial, tan_targeted_2022, dong2025safe, schumann2025realistic} are a concern, and STEP is currently the only framework that allows users to test their models (and potential countermeasures) against such attacks out-of-the-box.

\section{STEP Framework}

\begin{figure*}
    \centering
    \includegraphics[width=\textwidth]{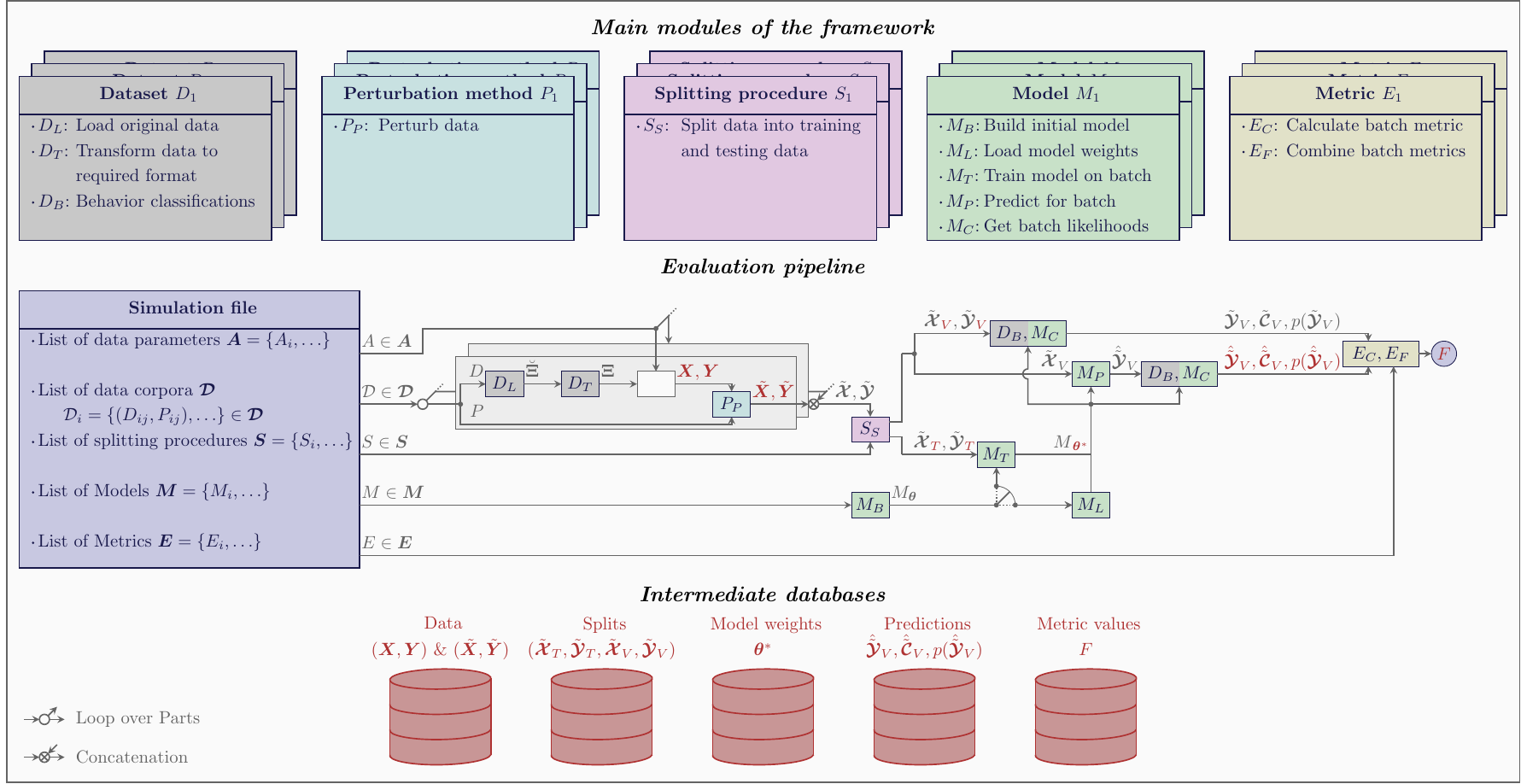}
    \caption{A high level overview of STEP (Structured Training and Evaluation Platform for benchmarking trajectory prediction models): 
    the five main modules (top row, see Section~\ref{sec:Modules}), the pipeline for training and evaluating models (middle row, see Section~\ref{sec:Experiemnt_running}), and the databases that the framework uses to save intermediate steps (bottom row). 
    In a simulation file, the user specifies the target data parameters, datasets, perturbation methods, splitting procedures, models, and metrics. The framework then follows the training/testing pipeline to calculate the performance value $F$ for each potential combination of these.
    }
    \label{fig:Framework_overview}
\end{figure*}

Our proposed framework STEP (Figure~\ref{fig:Framework_overview}), developed as a generalization of a previously proposed gap acceptance benchmarking framework~\cite{schumann_benchmarking_2023}, is designed to fill in the gaps in the existing literature (Table~\ref{tab:frameworkComparison}), providing 
\begin{itemize}
    \item a unified dataset format
    \item control over the data partitioning
    \item control over the observation length and frequency
    \item access to all relevant agent types and their features
    \item semantic scene information through images and scene graphs
    \item support for scene-level models during training and testing
    \item robustness testing 
    \item easy community-based expansion of all relevant modules (datasets, perturbation method, splitting procedure, models, and metrics) with comprehensive documentation.
\end{itemize}

There are two main use cases for our framework. First, users can easily integrate their own models, metrics, or datasets in STEP; our framework is highly modular so that can be achieved simply via \textit{adding a new module}. Second, the framework streamlines \textit{execution of experiments}. Users can run fully automated (yet flexible and easily customizable) training and testing of target models on target datasets, as well as calculation of the target metrics.

\subsection{Adding new modules} \label{sec:Modules}

\textbf{Datasets} are the fundamental building block for our framework, providing the scenarios in which behavior prediction models are trained and evaluated.
Currently, our framework includes 12 state-of-the-art datasets, ranging from the large-scale general datasets (\emph{Argoverse2}~\cite{wilson2021argoverse}, \emph{Lyft Level5}~\cite{houston2021one}, \emph{Waymo}~\cite{ettinger2021large}, \emph{NuScenes}~\cite{caesar_nuscenes_2020}, \emph{INTERACTION}~\cite{interactiondataset}) to those more focused on either specific driving contexts (roundabouts in \emph{rounD}~\cite{krajewski_round_2020}, highways in \emph{highD}~\cite{krajewski_highd_2018}, or intersections in \emph{inD}~\cite{bock2020ind}) or specific agent types such as pedestrians (\emph{ETH/UCY}~\cite{pellegrini2009you,lerner_crowds_2007}). To the best of our knowledge, STEP contains the most comprehensive library of datasets available, which includes all datasets that are part of the previous frameworks (Table~\ref{tab:frameworkComparison}). 

To add a new dataset to the Framework, a user has to implement two main functionalities as well as one optional functionality (cf. the pipeline in Fig.~\ref{fig:Framework_overview}):
\begin{itemize}
    \item $D_L$ to load the dataset's information $\breve{\bm{\Xi}}$ in its original form. In its most basic form, this will be information on the agents' states (positions, velocities, etc.), but can also include their type (vehicle, pedestrian, etc.) or size, as well as their environment, either based on top-down view images or graphs containing information such as lane positions.
    \item $D_T$ to transforms the original data into the framework-compliant format (i.e., $\bm{\Xi}$).
    \item In case the dataset supports high-level behavior classification (e.g., drivers approaching roundabouts making a discrete decision whether to yield to an oncoming car): $D_B$ to assign a provided scene (i.e., arbitrary observation $\bm{p}$, which might also be only the future trajectory $\bm{y}$) a high-level behavior classification $\bm{c}$. 
\end{itemize}

\textbf{Perturbation methods} can be applied to datasets to test robustness of trajectory prediction models, both against natural variability in the data and deliberate attacks ~\cite{hagenus2024survey}. The framework provides two perturbation methods based on state-of-the-art adversarial attack approaches, focused on perturbing either predicted vehicle's positions~\cite{zhang_adversarial_2022} or control states~\cite{cao_advdo_2022, schumann2025realistic}. Users can also add their own perturbation methods by implementing a function $P_P$ which takes a set of past observations $\bm{X}$ (including both agent's states and ancillary information) and corresponding future observations $\bm{Y}$ and perturbs them.

\textbf{Splitting procedures} split the dataset into the training and testing part. The framework provides a range of out-of-the-box splitting procedures:
\begin{itemize}
    \item Random splitting
    \item Splitting by dataset and/or locations, to allow testing of models on a location outside the training domain
    \item Random cross validation splitting, to test the influence of the splitting on the final performance metric.
    \item Splitting by safety critically to test models on the most important scenarios.
    \item Predefined splitting to fit to public challenges, such as on the \emph{Waymo} dataset.
\end{itemize}
Users can implement their own splitting methods by adding a module implementing the function $S_S$.

\textbf{Models} include implementations of prediction methods. This can be both trajectory prediction models (i.e., a conditional probability distribution $P(\bm{y} \mid \bm{x}, \bm{\theta})$ over future trajectories $\bm{y}$) or -- similar to~\cite{schumann_benchmarking_2023} -- behavior classification models (i.e., a distribution $P(\bm{c}\mid \bm{x}, \bm{\theta})$ over future behavior classes $\bm{c}$). The framework has out-of-the-box implementations for a variety of state-of-the-art trajectory prediction models (\emph{ADAPT}~\cite{aydemir_adapt_2023}, \emph{AgentFormer}~\cite{yuan_agentformer_2021}, \emph{AutoBot}~\cite{girgis_latent_2022}, \emph{FJMP}~\cite{rowe_fjmp_2023}, \emph{FloMo}~\cite{scholler_flomo_2021}, \emph{MTR}~\cite{shi_motion_2022}, \emph{Trajectron++}~\cite{salzmann_trajectron_2020}, \emph{TrajFlow}~\cite{meszaros_trajflow_2024}, \emph{Wayformer}~\cite{nayakanti_wayformer_2023}) as well as several behavior classification models~\cite{schumann_using_2023, theofilatos_cross_2021, xie_data-driven_2019}.

To integrate a new model to the framework, the following functions should be implemented:
\begin{itemize}
    \item $M_B$ to initialize the model with random parameters.
    \item $M_T$ to adjust the model parameters $\bm{\theta}$ to better fit the presented training data (passed in batches) and save the final parameters. The framework allows to automatically split a subset from the training set for validation, but as this is model-dependent, this has to be implemented for each model separately.
    \item $M_L$ to load and assign previously saved model parameters $\bm{\theta}^*$ to the model.
    \item $M_P$ to take the past observations $\bm{x}_V$ from the test set and sample a number of potential future trajectories for trajectory prediction models ($\hat{\bm{y}}_V \sim P(\bm{y} \mid \bm{x}_V, \bm{\theta}^*)$) or the predicted behavior probabilities $\hat{\bm{c}}_V \sim P(\bm{c}\mid \bm{x}, \bm{\theta})$.
    \item (optional) $M_C$ to calculate the likelihoods $p(\bm{y}) = P(\bm{y} \mid \bm{x}_V, \bm{\theta}^*)$ of a given future trajectory $\bm{y}$ according to the trained model and given the past observations $\bm{x}$. For models with explicitly defined probability function, it is recommended that users specify $M_C$. This is, however, optional, as the framework can estimate the likelihoods of sampled trajectories using robust multi-modal density estimation~\cite{meszaros2024rome}. 
\end{itemize}

\textbf{Metrics} assign a model a single scalar value given a specific combination of dataset and splitting method, and are generated automatically to facilitate comprehensive model comparison. STEP's large library of metrics covers all metrics commonly used in the recent literature, ranging from simple distance metrics such as average displacement error (ADE) to likelihood based ones like the negative log likelihood (NLL) or behavior based ones such as area under curve (AUC). An up to date list of metrics can be found in the documentation.

To include a new metric in the framework, a user need to implement two functions:
\begin{itemize}
    \item $E_C$ to take the ground truth data (trajectories $\bm{y}$, behavior classes $\bm{c}$, or trajectory likelihoods $p(\bm{y})$) batch-wise and compare it to the corresponding model predictions, resulting in a batch metric $f$.
    \item $E_F$ to combine the batch-wise values $f$ into a combined value $F$. 
\end{itemize}

\subsection{Running an experiment} \label{sec:Experiemnt_running}
To set up an experiment, users need to create a \textit{simulation file} that specifies which modules are used in the experiment.
\begin{itemize}
    \item \textbf{Data corpus} $\mathcal{D} = \{(D_1, P_1), \hdots\}$ that assembles data from multiple \textit{datasets} $D$, where each dataset can be optionally paired with a \textit{perturbation method} $P$ (which could be empty in case the data is to remain unperturbed).
    \item \textbf{Data parameter setting} $A = \{ n_I, n_O, \delta t\}$ to set the number of input (past observed) and output (future) time steps, and the time step value (in seconds). These are used consistently across different $\mathcal{D}$, facilitating apples-to-apples comparisons across datasets.
    \item \textbf{Splitting method} $S$ specifying the size and characteristic (e.g., the test location(s) when splitting based on scenario locations), or the inclusion/exclusion of (potentially perturbed) data in training or testing set.
    \item \textbf{Model} $M$ to be trained/tested; this can include the setting of specific model hyperparameters, allowing the comparison of different model variants.
    \item \textbf{Metric} $E$ determining the formulas used for the calculation of the final performance values $F$.
\end{itemize}

In case the simulation file includes multiple modules of a given type (e.g., multiple datasets, models, and metrics), the framework will evaluate each possible combination, provided that those are feasible (e.g., splitting the data into training and testing set based on dataset requires that the used data corpus $\mathcal{D}$ contains at least two datasets).
These final performance metrics $F$ are then determined by the framework in a multi-step process (Figure~\ref{fig:Framework_overview}).

While our framework allows the addition of both trajectory prediction and behavior classification models predicting the probabilities with which an agent (or group of agents) will perform certain classifiable maneuvers (such as which of two agents will cross an intersection first), for simplicity's sake, the following description of the evaluation pipeline is focused on the former.

As the first step, for each dataset module $D$ included in data corpus $\mathcal{D}$, the underlying raw data $\breve{\bm{\Xi}}$ is loaded and transformed into the internal format of our framework $\bm{\Xi}$ (detailed description provided in the online documentation) using the functions $D_L$ and $D_T$, respectively. From each scenario in this dataset -- based on the dataset parameters $A$ -- the framework extracts one or more samples (differing in the prediction time point $t_0$) with the respective input and output data ($\bm{X}$ and $\bm{Y}$). In the framework, one can set different methods for determining $t_0$, such as (1) using the first available point in time with sufficient past trajectories available, (2) a specific point in time before certain behaviors are initialized/completed (see~\cite{schumann_using_2023} for more details), or (3) regular spacing throughout the given scenario.
Importantly, while $\bm{Y}$ only contains the future (i.e., $t > t_0$) trajectories, $\bm{X}$ can contain not only the past observed trajectories but also additional predictive information such as agent type and size, or environment information (in image and/or graph format). The perturbation method is then applied to the extracted data (via the function $P_P$), resulting in the final perturbed values $\tilde{\bm{X}}$ and $\tilde{\bm{Y}}$ for the dataset module. After iterating over all pairwise combinations of $D$ and $P$, the resulting (potentially perturbed) datasets (with input $\tilde{\bm{X}}$ and output $\tilde{\bm{Y}}$) are then concatenated into the combined input $\tilde{\bm{\mathcal{X}}}$ and output $\tilde{\bm{\mathcal{Y}}}$. 

In the next step, the combined dataset is then split based on the chosen splitting method (function $S_S$), resulting in training set ($\tilde{\bm{\mathcal{X}}}_T$ and $\tilde{\bm{\mathcal{Y}}}_T$) and testing set ($\tilde{\bm{\mathcal{X}}}_V$ and $\tilde{\bm{\mathcal{Y}}}_V$). The former is then repeatedly split into batches, which are fed to the initialized (function $M_B$) model's training function $M_T$, resulting in a trained model.

Afterwards, in a similar manner, the testing set is split into batches as well. The input data $\bm{x}_V$ in each batch is presented to the trained model, resulting in predicted trajectories $\hat{\bm{y}}_V$. For the datasets which include information on high-level behavior classification, the framework extracts the high-level behaviors ($\hat{\bm{c}}_V$ and $\bm{c}_V$) using the function $D_B$; this is done both for the predicted ($\hat{\bm{y}}_V$) and the ground truth ($\bm{y}_V$) trajectories. Importantly, on this step $M_C$ calculates the likelihoods of those trajectories according to the underlying predicted trajectory distribution. These are then fed to the batch-wise-evaluation function $E_C$, resulting in the batch metric values $f$. Combining those over the whole testing set using the function $E_F$ results in the final performance metric value $F$.

Whenever the framework saves the results of some intermediate steps, it will preemptively check for the existence of those results and if so, load the existing data instead. Due to the vide variety in model structures, loading model weights requires the model-specific function $M_L$.

Additionally, while the framework will attempt to calculate a performance metric value for every combination of five main building blocks, this might not always be possible. For example, a metric based on high-level behaviors will only be calculated if the testing set contains at least one scenario coming from a dataset for which behavior extraction is defined (see function $D_B$).

Finally, on top of the basic training/testing pipeline, the framework supports additional processes such as fine-tuning already trained models on new datasets.

\section{Experiments}
To demonstrate the flexibility of the proposed framework, we conducted a number of experiments. In these experiments, we used four state-of-the-art models (ADAPT~\cite{aydemir_adapt_2023}, FJMP~\cite{rowe_fjmp_2023}, Autobot~\cite{girgis_latent_2022}, Wayformer~\cite{nayakanti_wayformer_2023}). We focus on Argoverse2~\cite{wilson2021argoverse} as the main large-scale dataset. Additionally we also use rounD~\cite{krajewski_round_2020} (a dataset focused on German roundabouts) to investigate behavior classification performance of the models. To this end, we extracted \textit{gap acceptance} scenarios~\cite{schumann_benchmarking_2023} from rounD, where a vehicle entering a roundabout can either yield to a vehicle inside the roundabout (reject the gap) or go first (accept the gap). 

As trajectory prediction metrics, we used mean ADE and mean FDE for single agents and joint distributions, as well as miss rate for single agents, all based on 6 randomly sampled trajectories. As the behavior classification metric we used area under curve (AUC) and the expected calibration error (ECE). Given that some models (e.g., ADAPT) only predict a limited number of trajectories (i.e., not samples from a distribution), we did not use metrics based on probabilities (BrierminFDE~\cite{wilson2021argoverse}, NLL~\cite{thiede2019analyzing}, etc.). 

\subsection{Evaluating model sensitivity to input data parameters}
Ideally, prediction models should be able to achieve good prediction performance with minimal observation horizons, to ensure an AV can react adequately even if a road user was just detected, or to ensure real-time computations. At the same time, longer observation horizons may contain subtle cues, which could improve prediction accuracy.
Consequently, we test the impact of different observation horizons on the final prediction performance, while also considering different observation frequencies (i.e. timestep length $\delta t$ of past observations), which vary considerably between different benchmarks~\cite{wilson2021argoverse, caesar_nuscenes_2020}.

Generally in its benchmarks, the Argoverse2 dataset uses $n_I = 50$ input timesteps and $n_O = 60$ output timesteps with a timestep length of $\delta t = 0.1s$. 
To evaluate the influence of the given input, we still predict the same $6s$, but now with a timestep length of $\delta t \in \{0.5s, 0.25s, 0.1s\}$. To ensure that under each of these conditions the first observation is identical (so one setting does not has access to more observations) the time $T_I = \delta t (n_I-1)$ from first observation to prediction time $t_0$ needs to be divisible by all tested $\delta t$. Therefore, we cannot use the common value of $T_I = \SI{4.9}{s}$, and instead use $T_I = \SI{4.5}{s}$ as our large observation horizon and $T_I = \SI{1.5}{s}$ as the small one.

\begin{table*}[]
    \centering
    \caption{Sensitivity of model performance to input data parameters (models trained on Argoverse2). In each cell, the rows correspond to the different timestep lengths $\delta t \in \{0.5s, 0.25s, 0.1s\}$, while the columns represent different observation horizons $T_I \in \{1.5s, 4.5s\}$. Hence, in each cell the upper left number corresponds to the shortest and least detailed input, while the the lower right number represents the longest, most detailed input. The model with the best value for each metric and input parameter combination is shown in bold. The best value achieved over the different input conditions by each model is underlined.}
    \label{tab:exp1}
    \begin{tabular}{|lr|cc|cc|cc|cc|cc|} 
    \toprule
     & Metric & \multicolumn{2}{c|}{$\text{minADE}_6 \downarrow$} & \multicolumn{2}{c|}{$\text{minADE}_{J,6} \downarrow$} & \multicolumn{2}{c|}{$\text{minFDE}_6 \downarrow$} & \multicolumn{2}{c|}{$\text{minFDE}_{J,6} \downarrow$} & \multicolumn{2}{c|}{$\text{MR}_6 \downarrow$} \\ 
    Model & $\delta t$ / $T_I$ & \SI{1.5}{s} & \SI{4.5}{s} & \SI{1.5}{s} & \SI{4.5}{s} & \SI{1.5}{s} & \SI{4.5}{s} & \SI{1.5}{s} & \SI{4.5}{s} & \SI{1.5}{s} & \SI{4.5}{s} \\ 
    \midrule
    \multirow{3}{*}{ADAPT~\cite{aydemir_adapt_2023}} & \SI{0.5}{s} & \textbf{0.780} & 0.882 & 2.353 & 2.278 & 1.764 & 1.734 & 5.389 & 4.685 & 0.153 & 0.136 \\ 
     & \SI{0.25}{s} & 0.824 & \underline{\textbf{0.730}} & 2.036 & \underline{1.752} & 1.916 & \underline{1.451} & 4.951 & \underline{4.386} & 0.164 & \underline{0.119} \\ 
     & \SI{0.1}{s} & 1.071 & \textbf{0.856} & 2.184 & 1.938 & 1.692 & 2.205 & 5.395 & 4.404 & 0.181 & 0.128 \\ 
    \midrule
    \multirow{3}{*}{FJMP~\cite{rowe_fjmp_2023}} & \SI{0.5}{s} & 0.849 & 0.982 & \textbf{1.732} & 1.980 & 1.803 & 1.842 & \textbf{3.772} & 4.487 & 0.202 & 0.213 \\ 
     & \SI{0.25}{s} & \underline{\textbf{0.743}} & 0.772 & \underline{\textbf{1.446}} & \textbf{1.620} & 1.696 & \underline{1.650} & \underline{\textbf{3.610}} & \textbf{3.733} & \underline{0.198} & 0.204 \\ 
     & \SI{0.1}{s} & 1.001 & 0.965 & 1.853 & \textbf{1.678} & 2.346 & 1.905 & 4.869 & 4.269 & 0.274 & 0.230 \\ 
    \midrule
    \multirow{3}{*}{AutoBot~\cite{girgis_latent_2022}} & \SI{0.5}{s} & \underline{1.308} & 1.931 & \underline{2.126} & 2.781 & \underline{2.113} & 2.601 & \underline{4.127} & 4.539 & \underline{0.208} & 0.344 \\ 
     & \SI{0.25}{s} & 1.484 & 2.318 & 2.255 & 3.211 & 2.333 & 2.920 & 4.315 & 4.834 & 0.251 & 0.417 \\ 
     & \SI{0.1}{s} & 1.614 & 1.628 & 2.401 & 2.461 & 2.578 & 2.498 & 4.655 & 4.672 & 0.280 & 0.277 \\ 
    \midrule
    \multirow{3}{*}{Wayformer~\cite{nayakanti_wayformer_2023}} & \SI{0.5}{s} & 0.893 & \underline{\textbf{0.826}} & 1.949 & \textbf{1.811} & \textbf{1.428} & \underline{\textbf{1.339}} & 4.084 & \textbf{3.872} & \textbf{0.091} & \underline{\textbf{0.085}} \\ 
     & \SI{0.25}{s} & 0.915 & 0.826 & 1.872 & 1.727 & \textbf{1.496} & \textbf{1.370} & 4.082 & \underline{3.866} & \textbf{0.094} & \textbf{0.087} \\ 
     & \SI{0.1}{s} & \textbf{0.941} & 0.881 & \textbf{1.762} & \underline{1.714} & \textbf{1.618} & \textbf{1.492} & \textbf{4.058} & \textbf{3.960} & \textbf{0.103} & \textbf{0.094} \\ 
    \bottomrule
    \end{tabular}
\end{table*}

\subsection{Evaluating the variability of the training process}
The performance of most machine learning models is stochastic due to random initialization of weights or the random order in which training samples are presented. However, in many benchmarks, this randomness is often not acknowledged. Therefore, we investigated variability in model performance by training the same model multiple times\footnote{In our evaluation of different timestep lengths $\delta t$ and observation horizons $T_I$, the best results were most often achieved for $\delta t = 0.25s$ and $T_I = 4.5s$ (Table~\ref{tab:exp1}). Consequently, in this and following experiments, we used these settings.}.

In addition to varying the random seed of the model, we replaced evaluation on the predefined train/test partition of Argoverse2 with nine-fold cross-validation. Specifically, the whole dataset (train and test set combined) was randomly divided into nine equally large parts (matching the size of the predefined testing set in Argoverse2, which includes approx. 11\% of all samples), each of which served as the test set in one of the nine training runs (with the other eight parts serving as the train set).

\subsection{Evaluating generalizability}
For a prediction model to be truly useful, it needs to be generalizable over different geographic locations. 
Therefore, we evaluated the performance of models trained on the Argoverse2 dataset (collected mostly in US) on a random subset of the rounD dataset (collected in Germany). 
We compared the resulting performance of the models to the same models but trained only on the remaining rounD data. 
Furthermore, we compared it to the models trained on Argoverse2 and then fine-tuned on rounD, using only a twentieth of the training epochs and 0.4 times the original learning rate compared to the model trained on rounD exclusively.

\subsection{Evaluating robustness}
Besides generalizability, another important aspect of model performance is robustness to input perturbations~\cite{hagenus2024survey}. To illustrate how our framework can be used to test model robustness, we generated perturbations using adversarial attacks against the models trained exclusively on rounD in the previous experiment, with varying limits on allowed perturbations $d_{\max} \in \{\SI{0.25}{m}, \SI{0.5}{m}, \SI{1}{m}\}$. Here, we used attacks on control actions with limits on the perturbed future trajectory to ensure that comparisons between the ground truth future trajectories and the predictions resulting from the perturbed data are meaningful~\cite{schumann2025realistic}. However, as these attacks rely on repeated calculations of the gradient through the target prediction model, we were unable to generate attacks on FJMP due to its intensive data preprocessing in numpy. All other models from the previous experiments were evaluated on respective perturbed datasets.

\section{Results}
\subsection{Sensitivity to input data parameters}
We found that out of the three tested $\delta t$ values, $\delta t = 0.1s$ generally has the poorest overall results (Table~\ref{tab:exp1}) for both tested observation horizons. For $T_I=4.5s$ specifically, only AutoBot achieves its best performance values across most of the metrics at $\delta t = 0.1s$, despite this setting being the closest to the standard setting for the Argoverse2 dataset. For all other models, the best results when using an input duration of $T_I=4.5s$ are generally achieved with the timestep length of $\delta t = 0.25s$.

In the case of an input duration of $T_I=1.5s$, the results are not as clear as with an input duration of $T_I=4.5s$. The best performance is generally achieved at $\delta t = 0.5s$, however similar performance is often achieved with the timestep length of $\delta t = 0.25s$ as well.

These results suggest that inputs with small $\delta t$ provide too detailed information to models, resulting in overfitting to minute differences and in turn a deterioration of performance. At the same time, there is no single combination of $\delta t$ and $T_I$ that consistently yields the best performance across models. In the case of an input duration of $T_I=1.5s$, the lack of clear difference between $\delta t = 0.25s$ and $\delta t = 0.5s$ can be attributed to the short input duration, where an input timestep length of $\delta t = 0.25s$ does not provide much more additional information compared to an input timestep length of $\delta t = 0.5s$, particularly in urban driving scenarios where behavior changes over a time period of $T_I = 1.5s$ are generally minor.

Overall, our experiments demonstrate that the interplay between timestep length and observation horizon can substantially affect model performance, but is not trivial and thus warrants further investigation. At the same time, our framework can aid researchers in analyzing the sensitivity of models to these parameters.

\subsection{Variability of the training process}

\begin{figure*}
    \centering
    \includegraphics{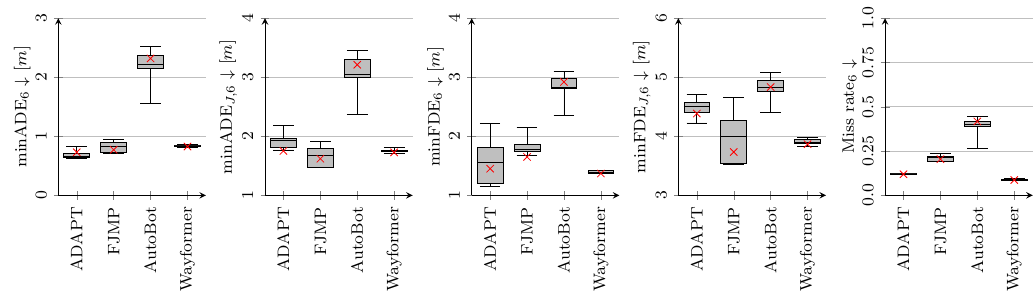}
    \caption{Variability of model performance due to variations in train/test split. Boxplots represent metric values resulting from the nine-fold cross validation on Argoverse2 with $\delta t = 0.25s$ and $T_I = 4.5s$. Red crosses show the values achieved under the same settings on the predefined test-set (Table~\ref{tab:exp1}).}
    \label{fig:Variety}
\end{figure*}
Our experiment demonstrated that random variations in training data can have a major effect on model performance (Figure~\ref{fig:Variety}). Hence, evaluating models' performance based on a single training run on a predefined test-set (which is a common practice in the field) can be misleading. 

For example, based on Table~\ref{tab:exp1} it would appear that FJMP has superior performance on the joint metrics. However, Figure~\ref{fig:Variety} shows that FJMP's performance in joint metrics is highly variable depending on the variations in the train/test split and can 
be better or worse compared, for instance, to the performance of Wayformer. 

Furthermore, some models exhibit higher levels of variability than others, with AutoBot especially showing consistently wide spreads in results, whereas Wayformer displays a consistently narrow spread.
The low variance of Wayformer indicate that the high performance variability of models such as AutoBot is a result of both the model structure and training process, as opposed to being caused only by the random splitting of the dataset.
As such, reporting the results for a instance of the model architectures -- as is common in the field -- can lead to a skewed perception of model performance. 
Reporting the results for a single instance of the model architectures is particularly an issue when the selection method for the used instances is not detailed.

Overall, our results suggest that the field would benefit from more transparent evaluation of model performance variability. Model developers can better illustrate the benefits of their models by training multiple model instances with varying random seeds and data splits -- something that our framework can streamline as well. Additionally, designing benchmarks in which there is only a single predefined testing set, might introduce an artificial bias in model evaluation.

\subsection{Generalizability}
\begin{table*}[]
    \centering
    \caption{Evaluation of generalizability of models tested on rounD. In each cell, the upper value is is the result for the model instance trained only on Argoverse2 (A), the middle value is the result for the model trained on Argoverse2 before being fine-tuned on a small subset of rounD (AR), and the lower value is the result for the model trained only on rounD (R). The model with the best value for each metric and training process is shown in bold, while the best value achieved over the different training process by each model is underlined.}
    \label{tab:exp3}
    \begin{tabular}{|lr|c|c|c|c|c|c|c|} 
    \toprule
    Model & Source & $\text{minADE}_6 \downarrow$ & $\text{minADE}_{J,6} \downarrow$ & $\text{minFDE}_6 \downarrow$ & $\text{minFDE}_{J,6} \downarrow$ & $\text{MR}_6 \downarrow$ & $\text{AUC} \uparrow$ & $\text{ECE} \downarrow$ \\ 
    \midrule
    \multirow{3}{*}{ADAPT~\cite{aydemir_adapt_2023}} & A & \textbf{3.059} & \textbf{4.075} & \textbf{6.342} & \textbf{10.030} & 0.694 & \textbf{0.984} & 0.420 \\
     & AR & 1.817 & 2.745 & \underline{3.589} & 6.732 & 0.536 & 0.984 & 0.500 \\
     & R & \underline{1.735} & \underline{2.582} & 3.619 & \underline{6.065} & \underline{0.509} & \underline{\textbf{0.990}} & \underline{0.419} \\
    \midrule
    \multirow{3}{*}{FJMP~\cite{rowe_fjmp_2023}} & A & 5.813 & 7.255 & 15.068 & 18.488 & 0.840 & 0.966 & \textbf{0.419} \\
     & AR & 2.608 & 3.224 & 6.988 & 8.874 & 0.699 & \textbf{0.990} & \textbf{0.340} \\
     & R & \underline{2.022} & \underline{2.452} & \underline{5.778} & \underline{7.156} & \underline{0.645} & \underline{\textbf{0.990}} & \underline{\textbf{0.259}} \\
    \midrule
    \multirow{3}{*}{AutoBot~\cite{girgis_latent_2022}} & A & 5.004 & 6.208 & 12.459 & 15.867 & 0.829 & 0.974 & 0.499 \\
     & AR & 1.876 & 2.337 & \underline{3.175} & \underline{\textbf{4.720}} & \underline{0.390} & \underline{0.989} & 0.499 \\
     & R & \underline{1.646} & \underline{2.138} & 3.989 & 5.341 & 0.442 & 0.986 & \underline{0.421} \\
    \midrule
    \multirow{3}{*}{Wayformer~\cite{nayakanti_wayformer_2023}} & A & 3.211 & 4.358 & 8.066 & 12.021 & \textbf{0.616} & 0.978 & 0.500 \\
     & AR & \underline{\textbf{1.272}} & \underline{\textbf{1.881}} & \underline{\textbf{2.682}} & \underline{4.828} & \underline{\textbf{0.191}} & \underline{0.988} & \underline{0.500} \\
     & R & \textbf{1.359} & \textbf{1.951} & \textbf{2.981} & \textbf{4.911} & \textbf{0.263} & 0.974 & 0.500 \\
    \bottomrule
    \end{tabular}
\end{table*}
We found that models trained only on Argoverse2 were substantially worse on all performance metrics compared to models trained on rounD as well as Argoverse2-trained models fine-tuned on rounD (Table~\ref{tab:exp3}). This is as expected, as the models trained only on Argoverse2 are dealing with a scenario previously underrepresented or nonexistent in the training data. However, our results show that pre-training on a large dataset like Argoverse2 and then fine-tuning on a small subset of the target dataset (rounD) can result in performance comparable (and in some cases even superior) to a model trained on rounD from scratch with orders of magnitude more computational resources. 

When comparing different models, we observed that for trajectory prediction metrics, Wayformer tends to outperforms other models in cases when the training data included (parts of) the rounD dataset. 
While FJMP was comparable to Wayformer when training and testing on Argoverse2 (Figure~\ref{fig:Variety}), it proved to generalize much worse to rounD. 

Meanwhile, for those models trained only on Argoverse2, the best performance is generally produced by ADAPT, which seems to be the most generalizable model in the absence of fine-tuning.

\subsection{Robustness}
\begin{table*}[]
    \centering
    \caption{A comparison of different models on rounD with different training procedures and different perturbation settings for the metric $\text{minADE}_6 \downarrow$. In each cell, the upper value is is the result for the model trained only on Argoverse2 (A), the middle value is the result for the model trained on Argoverse2 before being finetuned on rounD (AR), and the lower value is the result for the model trained only on rounD (R). Meanwhile, the columns display the different attacked models and the maximum allowed perturbation of each attack. The model with the best value for each attack is shown in bold.}
    \label{tab:exp4}
    \begin{tabular}{|lr|c|ccc|ccc|ccc|} 
    \toprule
    \multicolumn{2}{|r|}{Perturbation} & \multirow{2}{*}{None} & \multicolumn{3}{c|}{ADAPT~\cite{aydemir_adapt_2023}} & \multicolumn{3}{c|}{AutoBot~\cite{girgis_latent_2022}} & \multicolumn{3}{c|}{Wayformer~\cite{nayakanti_wayformer_2023}} \\ 
    Model & Source / $d_{\max}$ &  & $\SI{0.25}{m}$ & $\SI{0.5}{m}$ & $\SI{1}{m}$ & $\SI{0.25}{m}$ & $\SI{0.5}{m}$ & $\SI{1}{m}$ & $\SI{0.25}{m}$ & $\SI{0.5}{m}$ & $\SI{1}{m}$ \\ 
    \midrule
    \multirow{3}{*}{ADAPT~\cite{aydemir_adapt_2023}} & A & {\textbf{3.059}} & \textbf{3.237} & \textbf{3.402} & \textbf{3.606} & \textbf{3.140} & \textbf{3.148} & \textbf{3.192} & \textbf{3.142} & \textbf{3.182} & \textbf{3.228} \\
    & AR & {1.817} & \textbf{1.922} & \textbf{2.097} & \textbf{2.357} & \textbf{1.819} & \textbf{1.819} & \textbf{1.885} & \textbf{1.831} & \textbf{1.831} & \textbf{1.881} \\
    & R &{1.735} & \textbf{1.821} & \textbf{1.991} & \textbf{2.319} & {\textbf{1.620}} & \textbf{1.725} & \textbf{1.794} & {\textbf{1.656}} & \textbf{1.727} & \textbf{1.827} \\
    \midrule
    \multirow{3}{*}{FJMP~\cite{rowe_fjmp_2023}} & A & 5.813 & 6.588 & 6.731 & 6.881 & 6.535 & 6.543 & 6.543 & 6.526 & 6.520 & 6.540 \\
     & AR & 2.608 & 12.077 & 12.047 & 12.029 & 12.048 & 12.082 & 12.056 & 12.037 & 12.071 & 12.047 \\
     & R &2.022 & 12.598 & 12.579 & 12.519 & 12.620 & 12.616 & 12.574 & 12.641 & 12.647 & 12.655 \\
    \midrule
    \multirow{3}{*}{AutoBot~\cite{girgis_latent_2022}} & A & 5.004 & 9.317 & 9.381 & 9.445 & 9.259 & 9.260 & 9.297 & 9.274 & 9.280 & 9.298 \\
     & AR & 1.876 & 12.276 & 12.285 & 12.287 & 12.270 & 12.274 & 12.278 & 12.271 & 12.269 & 12.275 \\
     & R &1.646 & 14.478 & 14.477 & 14.494 & 14.477 & 14.474 & 14.494 & 14.481 & 14.479 & 14.492 \\
    \midrule
    \multirow{3}{*}{Wayformer~\cite{nayakanti_wayformer_2023}} & A & 3.211 & 8.271 & 8.432 & 8.528 & 8.234 & 8.233 & 8.261 & 8.248 & 8.293 & 8.325 \\
     & AR & \textbf{1.272} & 9.357 & 9.457 & 9.613 & 9.327 & 9.333 & 9.399 & 9.346 & 9.497 & 9.639 \\
     & R &\textbf{1.359} & 10.206 & 10.238 & 10.309 & 10.190 & 10.172 & 10.163 & 10.178 & 10.160 & 10.228 \\
    \bottomrule
    \end{tabular}
\end{table*}

Lastly, we tested the four models' robustness against adversarial attacks, with the results shown in Table~\ref{tab:exp4}. 
On the whole, model performance generally deteriorates when perturbations are added to the data; this is to be expected of an adversarial attack.
There are two main factors that can help mitigate the effect of adversarial attacks based on the obtained results; increased size and diversity of the training set, and the model's architecture.
In the case of FJMP, Autobot and Wayformer, models trained on the larger, more diverse Argoverse2 dataset exhibit a lower degree of performance degradation than those trained only on the smaller, roundabout focused rounD dataset. 
While these models had previously been the worst-performing ones in the unperturbed case, the lower degree of performance deterioration in the case of perturbations could be explained by the increased size and diversity of the Argoverse2 dataset.

Nevertheless, the training data does not seem to be the only factor affecting a model's robustness, as illustrated by the ADAPT model.
When attacked with perturbations optimized against itself, ADAPT’s performance decrease is roughly 34\% when allowing perturbations $d_{\max} = \SI{1}{m}$, compared to the more than 6 to 7 fold increase in displacement error for Autobot and Wayformer respectively. However, we can generally observe that larger allowed perturbations when attacking ADAPT will lead to worse performance of the same model.
Interestingly, when the perturbation are generated by attacks on other models, the performance of ADAPT stays fairly comparable to that on the unperturbed data (especially for displacements $d_{\max} \leq \SI{0.5}{m}$), and even improves in some cases.
This could be due to the fact that as part of the perturbation, the trajectories are smoothed to ensure reasonable curvatures and accelerations. 
Combined with fairly small perturbations, this smoothing could positively contribute to the final performance of ADAPT. 
Overall ADAPT's architecture seems to innately be more robust compared to other methods, as was already observed previously when testing against distribution shifts. 
A deeper analysis of the model aspects as well as how the structure of the training datasets contribute to a model's robustness are valuable lines for future research.

Compared to ADAPT, the other models seem to be much more sensitive even to small perturbations, with attacks performed on different models still being similarly successful overall, while the allowed magnitude in displacements only had a minor influence. This highlights a potential vulnerability: evidently, adversarial agents would not need to have access to the specific model an AV might be using to perform a potentially dangerous attack against, as tuning the attack to another model trained on similar data might be sufficient to generate successful attacks.

\subsection{Comparing model performance}
Across the four experiments, no single model could consistently outperform the other models, not even on a single metric.
For example, while Wayformer generally achieved better performance in the input sensitivity tests (Table~\ref{tab:exp1}), exhibited low variability in the presence of random initializations (Figure~\ref{fig:Variety}), and reasonable generalizability particularly when fine-tuning on the target dataset (Table~\ref{tab:exp3}) it showed very poor robustness to perturbations (Table~\ref{tab:exp4}).
Conversely, ADAPT -- while not always achieving the best results in other experiments -- demonstrated superior robustness.

Similarly, differences in model performance could also be observed when comparing trajectory-based and behavior classification metrics in the generalizability experiment on rounD (Table~\ref{tab:exp3}). There, despite its poor performance on the trajectory-based metrics, FJMP achieved the best results in terms of behavior classification metrics (AUC and ECE), suggesting that the scene-wise prediction approach of FJMP allows for a superior prediction of high-level interactions (such as the gap acceptance behavior in rounD) compared to usually superior single-agent predictors such as WayFormer. 

\section{Conclusion}
With this work, we introduced STEP, our framework for benchmarking trajectory prediction models. We showed that the framework is able to apply the same model to different datasets, provide far-reaching control over the actual structure of the dataset, and is able to create challenging testing regimes using perturbations such as adversarial attacks. Based on those experiments, we found that more and denser input data does not necessarily correspond to better performance, even for the largest state-of-the-art models. Additionally, minute differences in model performance are often not statistically significant, given the noise introduced by the model training process. Furthermore, most models are not directly transferable to scenarios outside their training domain, but can produce acceptable results with simple finetuning. And lastly, especially large prediction models are very susceptible to adversarial attacks, even if those are performed against a different model.

Our findings indicate that a more holistic evaluation of models -- such as the one we have performed here with STEP -- is needed to determine their applicability in real-world tasks. Furthermore, while state-of-the-art models are achieving rather good performance under the standard setup of training and testing on given datasets, there remains room for future contributions on matters related to generalizability and robustness.

\bibliographystyle{jabbrv_ieeetr}
\bibliography{IEEEabrv,zotero_library,biblio}

\end{document}